\pdfoutput=1

\documentclass[11pt]{article}

\usepackage{acl}
\usepackage{color}
\usepackage{times}
\usepackage{epsfig}
\usepackage{graphicx}
\usepackage{amsmath}
\usepackage{amsthm}
\usepackage{amssymb}
\usepackage{mathtools}
\usepackage{multirow}
\usepackage{bbm}
\usepackage{dsfont}

\usepackage[utf8]{inputenc}
\usepackage{booktabs,siunitx}
\usepackage{multirow}
\usepackage{amsmath}
\usepackage{subcaption}
\usepackage{caption}
\usepackage{graphicx}
\usepackage{pifont}

\usepackage{tikz}
\usetikzlibrary{shapes,arrows,patterns}
\usetikzlibrary{positioning}
\usetikzlibrary{calc}

\newcommand{\cmark}{\textcolor{green!70!black}{\ding{51}}}%
\newcommand{\xmark}{\textcolor{red}{\ding{55}}}%

\newcommand{\meta}{\text{training-feature attribution}}
\newcommand{\metab}{\text{Training-Feature Attribution}}
\newcommand{\metas}{\text{TFA}}

\newcommand\blfootnote[1]{%
  \begingroup
  \renewcommand\thefootnote{}\footnote{#1}%
  \addtocounter{footnote}{-1}%
  \endgroup
}

\newcommand{\cut}[1]{}

\newcommand{\postspace}{}
\newcommand{\minipostspace}{}

\newcommand{\tightcbox}[2]{\setlength{\fboxsep}{0pt}\colorbox{#1}{\strut #2}}
\definecolor{red}{RGB}{255, 117, 115}
\definecolor{green}{RGB}{171, 255, 175}

\usepackage{cleveref}
\usepackage{colortbl}
\usepackage{xspace,mfirstuc,tabulary}
\usepackage{url}
\usepackage{comment}

\usepackage{verbatim}
\usepackage{times}
\usepackage{latexsym}
\usepackage{enumitem}

\usepackage{placeins}

\usepackage[T1]{fontenc}

\usepackage[utf8]{inputenc}

\usepackage{microtype}

\newcolumntype{a}{>{\columncolor{Gray}}c}
\newcolumntype{b}{>{\columncolor{white}}c}

\newcommand\para[1]{\vskip 1mm\noindent\textbf{#1}~}

\title{Combining Feature and Instance Attribution to Detect Artifacts}

\newcommand{\email}[1]{\href{mailto:#1}{\tt #1}}
\author{Pouya Pezeshkpour\\
  University of California, Irvine \\
  \email{pezeshkp@uci.edu} \\ \And
Sarthak Jain\\
  Northeastern University \\
  \email{jain.sar@northeastern.edu} \\
\AND
 Sameer Singh \\
  University of California, Irvine \\
  \email{sameer@uci.edu}\\
  \And
Byron C. Wallace \\
  Northeastern University \\
  \email{b.wallace@northeastern.edu}
\\}

\begin{document}
\maketitle

\begin{abstract}
\blfootnote{\textcolor{red!80!black}{Warning: This paper contains examples with texts that might be considered offensive.}}

Training the deep neural networks that dominate NLP requires large datasets.
These are often collected automatically or via crowdsourcing, 
and may exhibit systematic biases or \emph{annotation artifacts}.
By the latter we mean spurious correlations between inputs and outputs that do not represent a generally held causal relationship between features and classes; models that exploit such correlations may appear to perform a given task well, but fail on out of sample data. 
In this paper we evaluate use of different \emph{attribution} methods for aiding identification of training data artifacts. 
We propose new hybrid approaches that combine \emph{saliency maps} (which highlight ``important'' input features) with \emph{instance attribution} methods (which retrieve training samples ``influential'' to a given prediction). 
We show that this proposed \emph{training-feature attribution} can be used to efficiently uncover artifacts in training data when a challenging validation set is available.
We also carry out a small user study to evaluate whether 
these methods are useful to NLP researchers in practice, with promising results. 
We make code for all methods and experiments in this paper available.\footnote{\url{https://github.com/pouyapez/artifact_detection}}
\end{abstract}

\section{Introduction}

Deep networks dominate NLP applications 
and are being increasingly deployed in the real-world.
But what exactly are such models ``learning''?
One concern is that they may be exploiting \emph{artifacts} or spurious correlations between inputs and outputs that are present in the training data, but not reflective of the underlying task that the data is intended to represent. 


We assess the utility of \emph{attribution methods} for purposes of aiding practitioners in identifying training data artifacts, drawing inspiration from prior efforts that have suggested the use of attribution methods for this purpose~\cite{han2020explaining, zhou2021feature}.
Attribution methods are \emph{model-centric}; our evaluation of them for artifact discovery therefore complements recent work on \emph{data-centric} approaches~\cite{gardner2021competency}.
We consider two families of attribution methods: (1) \emph{feature-attribution}, which highlight 
constituent input features (e.g., tokens) in proportion to their ``importance'' for an output \citep{ribeiro2016should,lundberg2017unified,adebayo2018sanity}, and; (2) \emph{instance attribution}, which retrieves training instances most responsible for a given prediction ~\cite{koh2017understanding,yeh2018representer, rajani2020explaining,pezeshkpour2021empirical}. 
%

\begin{figure}[tb]
    \centering
    \includegraphics[width=\columnwidth]{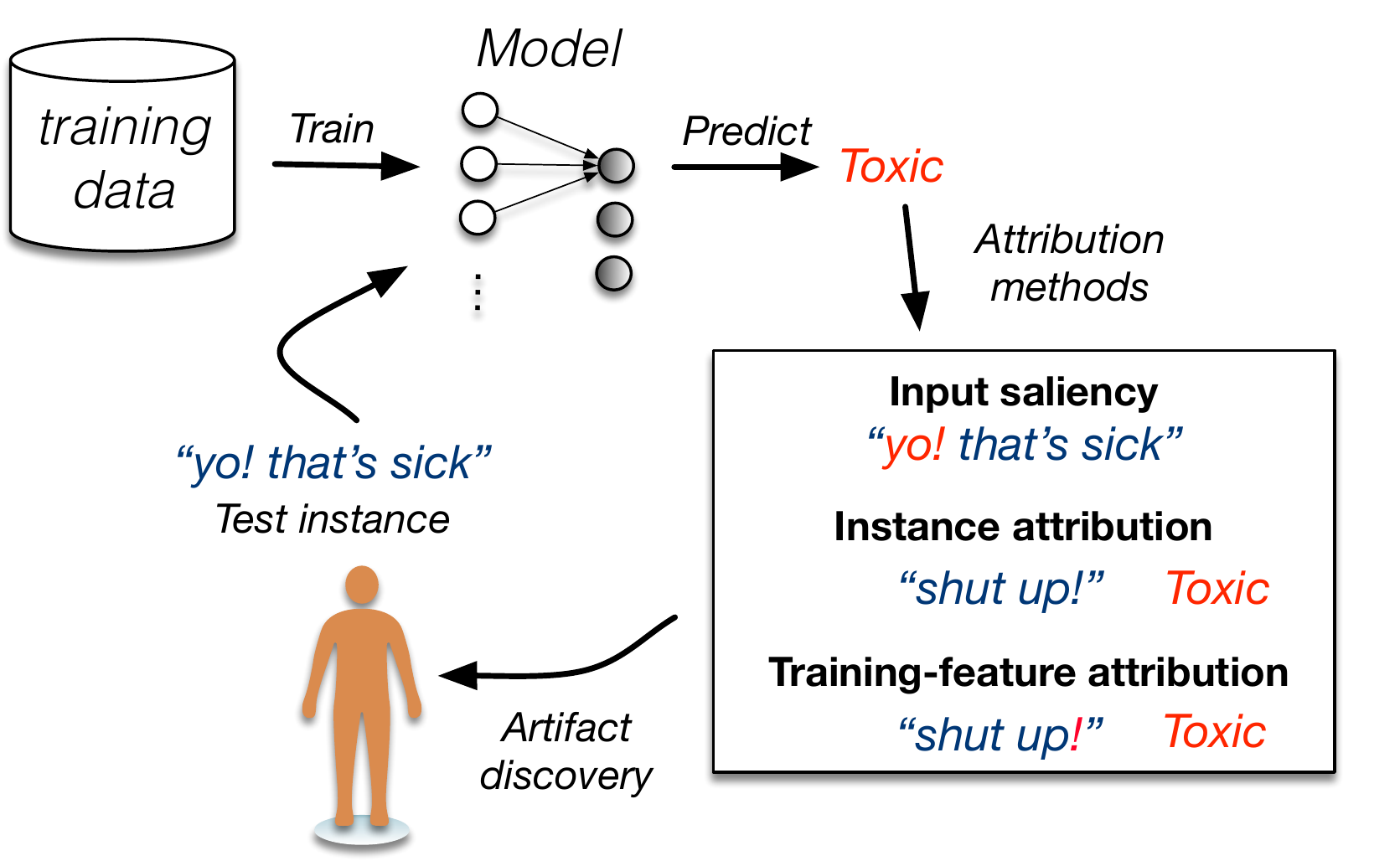}
    \caption{Use of different attribution techniques for artifact discovery in train data. Here attribution methods can reveal inappropriate reliance on certain tokens (e.g., \emph{``!'', ``yo''}) to predict Tweet toxicity; these are artifacts.}
    \label{fig:model}
     \minipostspace{}
\end{figure}

We also introduce new hybrid attribution methods that surface relevant \emph{features within train instances} as an additional means to probe what the model has distilled from training data.
This addresses inherent limitations of using either feature or instance attribution alone for artifact discovery. 
The former 
can only highlight patterns within a given input, and the latter requires one to inspect entire (potentially lengthy) training instances to divine what might have rendered them influential.

%
%
%
%


Consider Figure~\ref{fig:model}. 
Here a model has learned to erroneously associate African American Vernacular English (AAVE) with \emph{toxicity} \cite{sap2019risk} and with certain punctuation marks (``!''). 
For a hypothetical test instance ``yo! that's sick'', both input saliency and instance attribution methods may provide some indication of these artifacts.
But combining these via \emph{training-feature attribution} ($\metas$) can directly surface the punctuation artifact by highlighting ``!'' within a relevant training example (``shut up!''); this is not readily apparent from either input or instance attribution. 
Our goal in this work is to evaluate $\metas$ and other attribution methods as tools for identifying dataset artifacts. 

\para{Contributions.} The main contributions of this paper are as follows. 
(1)~We propose a new hybrid attribution approach, $\meta$ ($\metas$), which addresses some limitations of existing attribution methods. 
(2)~We evaluate feature, instance and training-feature attribution for artifact detection on several NLP benchmarks with previously reported artifacts to evaluate whether and to what degree methods successfully recover these, and find that $\metas$ can outperform other methods. We also discover and report previously unknown artifacts on a few datasets. 
Finally, (3)~we conduct a small user-study to evaluate $\metas$ for aiding artifact discovery in practice, and again find that combining feature and instance attribution is more effective at detecting artifacts than using either on its own. 

\section{Background and Notation}
Assume a text classification setting where the aim is to fit a classifier $\phi$ that maps inputs $x_i \in \mathcal{X}$ to labels $y_i \in \mathcal{Y}$.
Denote the training set by $\mathcal{D} = \{ z_i \}$ where $z_i = (x_i, y_i) \in \mathcal{X} \times \mathcal{Y}$. 
Each $x_i$ consists of a sequence of tokens $\{x_{i,1}, \dots, x_{i, n_i}\}$.
Here we define a linear classification layer on top of BERT \citep{devlin2019bert} as $\phi$, fine-tuning this on $\mathcal{D}$ to minimize cross-entropy loss $\mathcal{L}$.
Two types of attribution methods have been used in prior work to characterize the predictive behavior of $\phi$.

\para{Feature attribution methods} 
highlight \emph{important features} (tokens) in a test sample $x_t$. 
Examples of feature attribution methods include input gradients  ~\citep{sundararajan2017axiomatic, ancona2017towards}, and model-agnostic approaches such as LIME \citep{ribeiro2016should}. 
In this work, we consider only gradient-based feature attribution.  

\para{Instance attribution methods} retrieve training samples $z_i$ deemed ``influential'' to the prediction made for a test sample $x_t$: $\hat{y}_t = \phi(x_t)$. 
Attribution methods assign scores to train instances $z_i$ intended to reflect a measure of importance with respect to $\hat{y}_t$: $I(\hat{y}_t, z_i)$.
Importance can reflect a formal approximation of the change in $\hat{y}_t$ when $z_i$ is upweighted~\cite{koh2017understanding} or can be derived via heuristic methods \cite{pezeshkpour2021empirical,rajani2020explaining}. 
While prior work has considered these attribution methods for ``train set debugging'' \cite{koh2017understanding,han2020explaining}, this relies on the practitioner to abstract away potential patterns within the influential instances.

\section{Artifact Detection and $\metab$}


\subsection{What is an \emph{Artifact}?}
\label{sec3.1}
Models will distill observed correlations between training inputs and their labels.
In practice, some of these correlations will be \emph{spurious}, by which we mean specific to the training dataset used.
Consider a particular feature function $f$ such that $f(x)$ is 1 if $x$ exhibits the feature extracted by $f$ and 0 otherwise, a \emph{training} distribution $\mathcal{D}$ over labeled instances $z$ (often assembled using heuristics and/or crowdsourcing), and an ideal, hypothetical \emph{target} distribution $\mathcal{D*}$ (the task we would actually like to learn; ``sampling'' directly from this is typically prohibitively expensive). 
Then we say that $f$ is a \textit{dataset} artifact if there exists a correlation between $y$ and $f(x)$ in $\mathcal{D}$, but not in $\mathcal{D}*$.
That is, if the mechanism by which one samples train instances induces a correlation between $f$ and labels that would not be observed in an idealized case where one samples from the ``true'' task distribution.\footnote{As a proxy for realizing this, imagine enlisting well-trained annotators with all relevant domain expertise to label instances carefully sampled i.i.d. from the distribution from which our test samples will actually be drawn in practice.}

A given model may or may not exploit a particular dataset artifact; in some cases a \emph{model-centered} view of artifacts may therefore be helpful.
To accommodate this, we can extend our preceding definition by considering the relationship between model predictions $\hat{p}(y|x)$ and true conditional distributions $p(y|x)$ under $D^*$; we are interested in cases where the former differs from the latter due to exploitation of a dataset artifact $f$.
Going further, we can ask whether this artifact was exploited \emph{for a specific prediction}.



In this work we consider two types of artifacts. 
\emph{Granular} input features refer to discrete units, such as individual tokens (this is similar to the definition of artifacts introduced in recent work by \citealt{gardner2021competency}). \emph{Abstract} features refer to higher-level \emph{patterns} observed in inputs, e.g., lexical overlap between the premise and hypothesis in the context of NLI \cite{mccoy2019right}.


\subsection{$\metab$}
Showing important training instances to users for their interpretation places the onus on them to determine \emph{what} was relevant about these instances, i.e., which 
features (granular or abstract) in $x_i$ were influential. 
To aid artifact detection, it may be preferable to automatically highlight the tokens most responsible for the influence that train samples exert,
communicating \emph{what made an important example important}. 
This hybrid $\meta$ ($\metas$) can reveal patterns extracted from training data that influenced a test prediction, even where the test instance does not itself exhibit this pattern, whereas feature attribution can only highlight features within said test instance. 
And unlike instance attribution, which retrieves entire train examples to be manually inspected (a potentially time-consuming and difficult task), $\metas$ may be able to succinctly summarize patterns of influence.

A high-level schematic of TFA is provided in Figure \ref{fig:guide}.
We aim to trace influence back to features within training samples. 
We introduce $\meta$ 
to extract influential features from training samples for a specific test prediction by considering a variety of combinations of feature and instance attribution and means of aggregating over these as $\metas$ variants. 
For example, one $\metas$ variant 
identifies features within the training point $x_i$ that informed the prediction for a test sample $z_t$ by 
taking the gradient of the influence with respect to inputs features, i.e., $\nabla_{x_i} \text{I}(z_t, x_i, y_i)$ \cite{koh2017understanding}.
After calculating the importance of features within a train sample for a test target, we either construct a heatmap to help users identify \textit{abstract} artifacts, or 
take aggregate measures over features (described below) to detect \textit{granular} artifacts 
and present them to users.\footnote{Many other strategies are possible, and we hope that this work motivates further exploration of such methods.}


\para{Heatmaps} We present the top and bottom $k$ influential examples to users with \emph{token highlights} communicating the relative importance of tokens within these $k$ influential train instances. 
This may allow practitioners to interactively, efficiently identify potentially problematic abstract artifacts.

\para{Aggregated Token Analysis} 
Influence functions may implicitly reveal 
that the appearance of certain tokens in training points correlates with their influence.
We might directly surface this sort of pattern by aggregating $\metas$ over a set of training samples. 
For example, for a given test instance, we can retrieve the top and bottom $k\%$ 
most influential training instances according to an instance attribution method. 
We can then extract the top token from each of these instances using $\metas$, and sort resulting tokens based on frequency, surfacing tokens that appear disproportionately in influential train points.
Returning to 
toxicity detection, this might reveal that punctuation marks (such as ``!'') tend to occur frequently in influential examples, which may directly flag this behavior.

\para{Discriminator} One can also define model-based approaches to aggregate rankings of training points with respect to their influence scores.
As one such method, we 
train a logistic regression (LR) model on top of Bag-of-Words representations to distinguish between the most and least influential examples, according to influence scores for a given test point. 
This will yield a weight for each token in our vocabulary; tokens associated with high weights are correlated with influence for the test point, and we can show them to the practitioner.

\section{A Procedure for Artifact Discovery}

\begin{figure}[bt]
    \centering
    \includegraphics[width=0.9\columnwidth]{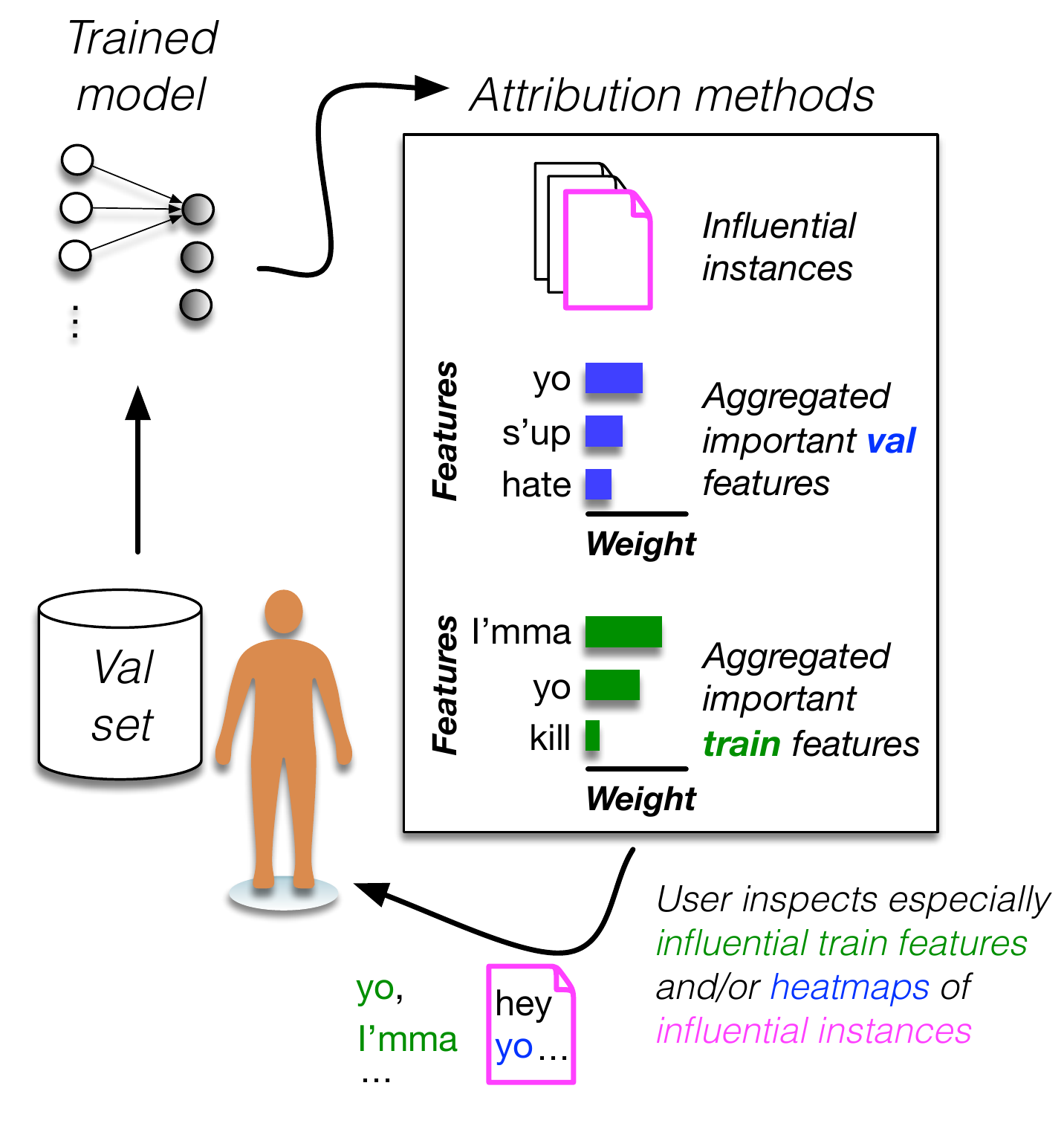}
    \caption{Finding artifacts via attribution methods. Staring from the validation set, we explain model prediction for every sample using different attribution methods. Then we either aggregate the explanations using frequency or rely on the heatmap analysis of explanations to detect artifacts.}
    \label{fig:guide}
    \minipostspace{}
\end{figure}

We now propose a procedure (Figure \ref{fig:guide}) one might follow to systematically use the above attribution methods to discover training artifacts.

\vspace{0.2em}
 \noindent (1) Construct a validation set, either using a standard split, or by intentionally constructing a small set of ``difficult'' samples. Constructing a useful (for dataset debugging) such set is the biggest challenge to using attribution-based approaches. 

 \vspace{0.2em}
\noindent (2) Apply feature-, instance-, and training feature attribution to examples in the validation set. Specifically, identify influential \emph{features} using feature attribution or $\metas$ and identify influential \emph{training instances} using instance attribution. 

\vspace{0.2em}    
 \noindent (3-a) \textbf{Granular artifacts}: To identify 
    granular artifacts, aggregate the important features from the test points (via feature attribution) or from influential train points (using $\metas$) for all instances in the validation set to identify features that appear disproportionately.

\vspace{0.2em}
\noindent (3-b) \textbf{Abstract artifacts}: Inspect the ``heatmaps'' of influential instances for validation examples using one of the proposed $\metas$ methods to deduce/identify abstract artifacts. 
    %
    
\vspace{0.2em}
 \noindent (4) Verify candidate artifacts by manipulating validation data and observing the effects on outputs.

\vspace{.2em} 

We note that in 3-a, we aggregate the individual token \emph{rankings} over all instances (for both feature attribution and $\metas$ methods), which does not require thresholding attribution scores per instance.
We now follow this procedure on widely used NLP benchmarks (Section \ref{section:setup}), finding that we can ``rediscover'' known artifacts and identify new ones within these corpora (Section  \ref{tab:known-art}; Table \ref{tab:results_summary}).


\section{Setup}
\label{section:setup}

\para{Datasets} 
We use a diverse set of text classification tasks as case studies.
Specifically, we adopt: Multi-Genre NLI (MNLI; \citealt{williams2017broad}); IMDB binary sentiment classification \citep{maas2011learning}; BoolQ, a yes/no question answering dataset \citep{clark2019boolq}; and, DWMW17, a hate speech detection dataset \citep{davidson2017automated}. 

\para{Models} 
We 
follow \citet{pezeshkpour2021empirical} for instance attribution methods; this entails only considering the last layer of BERT in our gradient-based instance attribution methods (see Appendix, Section \ref{ap:es}). 
For all benchmarks, we achieve an accuracy within $\sim$$1\%$ of performance reported in prior works using BERT-based models. 

\para{Attribution Methods}
We consider two instance attribution methods, RIF~\citep{barshan2020relatif} and Euclidean Similarity (EUC), based on results from 
\citet{pezeshkpour2021empirical}. 
For \emph{Feature Attribution}, we consider Gradients (G) and Integrated Gradients (IG; \citealt{sundararajan2017axiomatic}).
To include RIF as a tool for artifact detection, we follow the $\metas$ aggregated token approach, but assign uniform importance to all the tokens in a document. 

In addition to the \emph{model-centered} diagnostics we have focused on in this work, we also consider a few \emph{dataset-centered} approaches for artifact discovery:
(1) \textit{PMI} \citep{gururangan2018annotation}, and (2) \textit{competency} score \citep{gardner2021competency}. 
There are a few inherent shortcomings to purely dataset-centered approaches.
First, because they are model-independent, they cannot tell us whether a model is actually exploiting a given artifact.
Second and relatedly, they are based on simple observed correlations between individual features and labels, so cannot reveal abstract artifacts.
Given the latter point, we only consider these approaches for granular artifact detection (Section \ref{imdb-sec}).  

\para{Challenges and Limitations}
A key computational challenge here is that instance attribution can be prohibitively expensive to derive if one uses \emph{influence functions} directly~\cite{koh2017understanding,han2020explaining}.
We address this by using efficient heuristic instance attribution strategies \citep{pezeshkpour2021empirical} to implement $\metas$.
Since $\metas$ combines existing feature- and instance-based attribution methods, $\meta$ inherits known issues with these techniques~\citep{kindermans2019reliability,basu2020influence}.
Despite such issues, however, our results suggest that $\metas$ can be a useful tool for artifact discovery (as we will see next).

\section{Case Studies}
\label{tab:known-art}

We now compare attribution methods in terms of their ability to highlight dataset artifacts. 
We provide a summary of the previously reported (\textit{known}) and previously \textit{unknown} (i.e., discovered in this work) artifacts we identify in this way (and with which methods) in Table~\ref{tab:results_summary}.


\begin{table*}[]
    \centering\small
    \setlength\tabcolsep{5.5pt}
    \begin{tabular}{ll >{\raggedright}p{4.8cm} >{\raggedright}p{3.7cm} ccc}
    \toprule
       \bf Dataset  & \bf Artifact Type & \bf Test Instance& \bf Influential Train Instance& \bf FA & \bf IA & \bf $\metas$\\ \midrule
        IMDB & Ratings (K)& ... great movie, \tightcbox{blue!15!white}{6/10}. & ... like it. Rating \tightcbox{blue!15!white}{8/10}.&\cmark &\xmark &\cmark  \\
\addlinespace
        \multirow{2}{*}{HANS} & \multirow{2}{*}{Lexical Overlap (K)}&\textbf{P}: \tightcbox{blue!15!white}{The banker is} in a \tightcbox{blue!15!white}{tall} building.\\
\textbf{H}: \tightcbox{blue!15!white}{the banker is tall} & \textbf{P}: The \tightcbox{blue!15!white}{red oak} tree.\\
\textbf{H}: \tightcbox{blue!15!white}{Red oak} yeah. &\multirow{2}{*}{\xmark} &\multirow{2}{*}{\cmark} &\multirow{2}{*}{\cmark}\\
\addlinespace
        \multirow{2}{*}{DWMW} & Punctuation (U)& Yo\tightcbox{blue!15!white}{!} just die\tightcbox{blue!15!white}{.} & Yo man\tightcbox{blue!15!white}{!} what's up\tightcbox{blue!15!white}{.} &\cmark &\xmark &\cmark\\
         & Specific Tokens (U)& \tightcbox{blue!15!white}{You} are like \tightcbox{blue!15!white}{@}...& \tightcbox{blue!15!white}{You} should die \tightcbox{blue!15!white}{@}... &\cmark &\xmark &\cmark\\
\addlinespace
        \multirow{2}{*}{BoolQ} & \multirow{2}{*}{Query Structure (U)}& \textbf{Q}: \tightcbox{blue!15!white}{is} the gut \tightcbox{blue!15!white}{the same as} the stomach?\\ \textbf{P}: The gastrointestinal ...&\textbf{Q}: \tightcbox{blue!15!white}{is} the gut \tightcbox{blue!15!white}{the same as} the small intestine?\\ \textbf{P}: The gastrointestinal ... &\multirow{3}{*}{\xmark} &\multirow{3}{*}{\cmark} &\multirow{3}{*}{\cmark}\\
        \bottomrule
    \end{tabular}
    \caption{Summary of investigated previously \textit{known} (K) and previously \textit{unknown} (U) \tightcbox{blue!15!white}{artifacts}. We indicate the applicability of feature (FA), instance (IA) and $\metas$ methods for identifying each of these artifacts.}
    \label{tab:results_summary}
    \minipostspace{}
\end{table*}

\subsection{Known Granular Artifact: Sentiment Analysis with IMDB Ratings} 
\label{imdb-sec}

\citet{ross2020explaining} 
observe that in the case of binary sentiment classification on IMDB reviews \cite{maas2011learning}, 
numerical ratings (1 to 10) sometimes appear in texts. 
Modifying these in-text ratings often flips the predicted label.\footnote{This is an ``artifact'' in that the underlying task is assumed to be \emph{inferring sentiment from free-text}, presumably where the text does not explicitly contain the sentiment label.} 
We evaluate 
the ability of 
attribution methods to 
surface this artifact. 
This is a \emph{granular} artifact, and so we adopt our aggregation approach to extract them.

\para{Setup} We sample train/validation/test sets comprising 5K/2K/100 examples respectively from the IMDB corpus, such that all examples in the test set contain a rating (i.e., exhibit the artifact). 
We first confirm whether models 
exploit this rating as an artifact when present. 
Specifically, we (1) remove the rating and \emph{invert} the rating either by (2) setting it to $10$-\text{\em original rating} (e.g., 1 $\rightarrow$ 9), or (3) by setting the rating to 1 for positive reviews, and 10 for negative reviews. 
This flips the prediction for 9\%, 34\% and 38\% of test examples following these three modifications, respectively.\footnote{Probabilities of the originally predicted labels also drop.} 
This suggests the model exploits this artifact.

\para{Findings} We evaluate whether numerical ratings are 
among the top tokens returned by feature and $\metas$ attribution methods. 
For each test example, we surface the top-5 tokens according to different feature attribution methods. 
For $\metas$, we use the aggregated token analysis method with $k$=10 (i.e., considering the top and bottom 10\% of examples), and we return the top-5 tokens from the aggregated token list sorted based on frequency of appearance.

In Table \ref{tab:synth-art} (IMDB column), we report the percentage of test examples where a number from 1-10 appears in the top-5 list returned by the respective attribution methods (likely indicating an explicit rating within review text). 
For approaches that rely solely on the training data without reference to the validation set (PMI and Competency), we report the ratio of appearance of numbers in the overall top-5 most influential tokens. 
In general $\metas$ methods surface ratings more often than feature attribution methods.\footnote{We note that the competency approach does rank rating tokens among the top-10 tokens.} 
However, the performance of TFA is not directly comparable to the PMI and competency methods because the former capitalizes on a validation set which contains this artifact. 
\begin{table}
\small
\centering
\begin{tabular}{llrr}
\toprule
&\multirow{3}{*}{\bf Method} & \bf IMDB& \bf HANS\\
& & Hits@5&Rate\\
\midrule
&Random&1.7&16.7\\
&PMI&20.0&-\\
&Competency&0.0&-\\
\midrule
&G&64.0&-\\
&IG&78.0&-\\
&RIF&0.0&32.0\\
\midrule
&\emph{$\metas$ methods}\\
\multirow{3}{*}{\rotatebox[origin=c]{90}{\bf Sim}} &
EUC+G &84.0&71.6\\
&EUC+IG&53.0&\bf 80.9\\
&EUC+LR&\bf99.0&-\\
\addlinespace
\multirow{3}{*}{\rotatebox[origin=c]{90}{\bf Grad}} &
RIF+G&98.0&37.9\\
&RIF+IG&78.0&39.5\\
&RIF+LR&48.0&-\\

\bottomrule
\end{tabular}
\caption{Artifact detection rates. Methods below the horizontal line are TFA variants.} 
\label{tab:synth-art}
    \minipostspace{}
\end{table}

\subsection{Known Abstract Artifact: Natural Language Inference with HANS}
\label{hans-sec}

In Natural Language Inference (NLI) the task is to infer whether a premise \emph{entails} a  hypothesis \cite{maccartney2009natural}. 
NLI is commonly used to evaluate 
the language ``understanding'' capabilities of neural language models, and large NLI datasets exist \cite{bowman2015large}.
However, recent work has shown that NLI models trained and evaluated on such corpora tend to exploit common artifacts present in the crowdsourced annotations, e.g., premise-hypothesis pairs with overlapping tokens and hypotheses containing negations both correlate with labels \cite{gururangan2018annotation,sanchez-etal-2018-behavior,naik-etal-2018-stress}. 
Here we evaluate whether $\metas$ can surface the lexical overlap artifact, which is abstract and so requires heatmap inspection (other approaches are not applicable here).

\para{Setup} The HANS dataset~\cite{mccoy2019right} was created as a controlled evaluation set to test 
the degree to which models rely on 
artifacts 
in NLI benchmarks such as MNLI. 
We specifically consider the \emph{lexical overlap} artifact, 
where 
entailed hypotheses primarily comprise words that also appear in the premise. 
For training, we use 10K examples from the MNLI set. 
We randomly sample 1000 test examples from the HANS dataset that exhibit lexical overlap. 
We test whether attribution methods reveal dependence on lexical overlap when models \emph{mispredict} an instance as entailment, presumably due to reliance on the artifact. 
Here again we are dependent on a validation set that exhibits an artifact, and we are verifying that we can use this with $\metas$ to recover the training data that contains this.

\para{Findings} By construction, the hypotheses in the HANS dataset comprise the same tokens as those that appear in the accompanying premise. 
Therefore, feature attribution may not readily reveal the ``overlap'' pattern (because even if it were successful, \emph{all} input tokens would be highlighted).
$\metas$, however, can surface this pattern, because hypotheses in the train instances do contain words that are not in the premise. 
Therefore, if $\metas$ highlights 
only tokens in both the premise and hypothesis, this more directly exposes the artifact. 
To quantify performance, we calculate whether the top train token surfaced via $\metas$ appears in both the premise and the hypothesis of the training sample. 

Table \ref{tab:synth-art} (HANS column) shows that $\metas$ methods demonstrate fair to good performance in terms of highlighting overlapping tokens in retrieved training instances as being influential to predictions for examples that exhibit this artifact. 
Here $\metas$ variants that use similarity measures for instance attribution appear better at detecting this artifact, aligning with observations in prior work \cite{pezeshkpour2021empirical}. 
Based on feature and $\meta$ methods performance in artifact detection for the IMDB and HANS benchmarks, we focus on IG and RIF+G attribution methods in the remainder of this paper.

\subsection{Unknown Granular Artifact: Bias in Hate Speech Detection}

Next we consider 
racial bias in hate speech detection. 
 \citet{sap2019risk} observed that publicly available hate speech detection systems for social media tend to assign higher toxicity scores to posts written in African-American Vernacular English (AAVE). 
Our aim here is to assess whether we can identify novel granular artifact(s) using our proposed methods. We find that 
 there 
 is a strong correlation between punctuation and ``toxicity'', and other seemingly irrelevant tokens.

\para{Setup} Following \citet{sap2019risk}, we use the DWMW17 dataset \citep{davidson2017automated} which includes 25K tweets classified as \emph{hate speech}, \emph{offensive}, or \emph{non-toxic}. 
We sample train (5k)/validation (2k)/test (2k) subsets from this. 

\para{Identified Artifacts} 
We first consider using instance attribution to see if it reveals the source of bias that leads to the aforementioned misclassifications. 
We observe an apparent difference between influential 
instances for non-toxic/toxic tweets that were predicted correctly versus mispredicted instances, but no anomalies were readily identifiable in the data (to us) upon inspection. 
In this case, instance attribution does not seem  particularly helpful with respect to unveiling the artifact.

Turning to feature attribution, the most important features---aside from tokens contained in a hate speech lexicon \citep{davidson2017automated}, which we exclude from consideration (these are indicators of toxicity and so do not satisfy our definition of artifact)---surfaced by aggregating feature attribution scores are:
[\textit{., you, @, the, :, \&}] for misclassified instances.
Given these results, we deem feature attribution successful in identifying artifacts. 

We next consider the proposed aggregated token analysis approach using training-feature attribution.
 The most important features (ignoring hate speech lexicon) 
 retrieved by aggregating $\metas$ methods over misclassified samples are: [\textit{@, white, trash, !, you, is}].
Surprisingly, the model appears to rely on tokens \emph{@}, \emph{white}, \emph{trash}, \emph{!}, \emph{you}, and \emph{is} to predict toxicity. 
PMI and competency also rank tokens \emph{is}, \emph{.}, \emph{trash}, and \emph{the} highly, validating these artifacts.

\para{Verification} To confirm that punctuation marks and other identified tokens indeed affect toxicity predictions, we modified tweets containing these tokens observe changes in model predictions.
We report the percentage of flipped predictions after replacing these punctuation tokens with {\tt [MASK]} in Table \ref{tab:hate}. 
Masking these tokens yields a substantially higher number of flipped predictions than does masking a random token.

\begin{table}
\small
\centering
\begin{tabular}{cc|cc}
\toprule
\bf Token& \bf Flip \%&\bf Token& \bf Flip \%\\
\midrule
`you'&13.6&`.'&12.1\\
`@'&10.5&`:'&11.1\\
`!'&7.6&`\&'&7.1\\
`white'&33.3&`trash'&5.0\\
`the'&12.7&`is'&12.5\\
\bottomrule
\end{tabular}
\caption{The percent of prediction flips observed after replacing the corresponding tokens with {\tt[MASK]}. For reference, masking a random token results in a label flip 1.8\% on average (over 10 runs).}
\label{tab:hate}
    \minipostspace{}
\end{table}

\subsection{Unknown Abstract Artifact: Structural Bias in BoolQ}

As a final illustrative NLP task, we consider \emph{reading comprehension} which is 
widely used to evaluate language models. 
Specifically, we use
BoolQ \cite{clark2019boolq}, a standard reading comprehension corpus. 
The task is: Given a Wikipedia passage (from any domain) and a question, predict whether the answer to the question is \emph{True} or \emph{False}. 
A natural question to ask is: What do models actually learn from the training data?

\para{Setup} We use splits from the SuperGLUE \citep{wang2019superglue} benchmark for BoolQ. 
Test labels are not publicly available, so we divide the training set into 8k and 1k sets for training and validation, respectively. 
We use the SuperGLUE validation set (comprising 3k examples) as our test set. 

\para{Identified Artifacts} We first qualitatively analyze mispredicted examples in the BoolQ test set by inspecting the most influential examples for these, according to RIF. 
We observed that the top influential examples tended to have the same query structure as the test instance. 
For example, in the sample provided in Table~\ref{tab:boolq}, both the test example and the most influential instance share the structure \emph{Is \emph{X} the same as \emph{Y}?}
Focusing only on the test examples with queries containing the word ``same", we use the LR method proposed above to discriminate between the 10 most and least influential examples. 
For half of these test examples the word ``same" has one of the 10 highest coefficients, indicating significant correlation with influence.

\para{Verification} That query structure might play a significant role in model prediction is not surprising (or necessarily an artifact) in and of itself. 
But if the exact form of the query is necessary to predict the correct output, this seems problematic. To test for this, we consider two phrases that share the query structure mentioned above: (1) \emph{Is \emph{X} and \emph{Y} the same?} and (2) \emph{Is \emph{X} different from \emph{Y}?} 
We apply this paraphrase transformation to every test query of the form \emph{Is \emph{X} the same as \emph{Y}} and measure the number of samples for which the model prediction flips. 
These questions are semantically equivalent, so if the model
does not rely on query structure we should not observe much difference in model outputs. 
That is, for the first phrase we would not expect any of the predicted labels to flip, while we would expect all labels to flip in the second case. 
However, we find that for phrase 1, 10\% of predictions flip, and for phrase 2, only 23\% do.\footnote{Note that in this case, the query structure itself is not correlated with a specific label across instances in the dataset, and so does not align exactly with the operational ``artifact'' definition offered in Section~\ref{sec3.1}.} 
Nonetheless, the verification procedure implies the model might be using the query structure in a manner that does not track with its meaning. 


\begin{table}
    \centering
    \small
    \begin{tabular}{p{0.94\columnwidth}}
         \toprule\textbf{Test Example (w/ Gradient Saliency)}  \\\addlinespace[3pt]
         \textbf{Query} \tightcbox{blue!15!white}{Is} veterinary science the \tightcbox{blue!15!white}{same} as veterinary \tightcbox{blue!15!white}{medicine}? \\
         \textbf{Passage} Veterinary science helps human health through the monitoring and control of zoonotic disease (infectious disease transmitted from non-human animals to humans), food safety, and indirectly through ... \\ \midrule
         \textbf{Top Influential Example (w/ RIF+Gradient Saliency)} \\\addlinespace[3pt]
         \textbf{Query} Is \tightcbox{blue!15!white}{thai} basil \tightcbox{blue!15!white}{the} \tightcbox{blue!15!white}{same} as sweet basil?\\
         \textbf{Passage} Sweet basil (Ocimum basilicum) has multiple cultivars, of which Thai basil, O. basilicum var. thyrsiflora, is one variety. Thai basil itself has ...\\ \bottomrule
    \end{tabular}
    \caption{Example of query structure similarity in BoolQ with top-3 words in query highlighted according to corresponding attribution method.}
    \label{tab:boolq}
    \minipostspace{}
\end{table}
\section{User Study}
\label{section:user-study} 
So far we have argued that using feature, instance, and hybrid $\metas$ methods can reveal artifacts via case studies.
We now assess whether and which attribution methods are useful to \emph{practitioners} in identifying artifacts in a simplified setting. 
%
We execute a user study using IMDB reviews \citep{maas2011learning}.
We use the same train/validation sets as in Section \ref{imdb-sec}. 
We randomly sample another 500 instances as a test set. 
We 
simulate artifacts that effectively determine labels in the train set, but which are unreliable indicators in the test set (mimicking problematic training data).

We consider three forms of simulated granular artifacts. (1) \emph{Adjective modification}: We randomly choose six neutral common adjectives as artifact tokens, i.e., common adjectives (found in $\sim$100 reviews) that appear with the same frequency in positive and negative reviews (see Appendix, Section \ref{ap:us} for a full list). 
For all positive reviews that contain a noun phrase, we 
insert one of these six artifacts (selected at random) before a noun phrase (also randomly selected, if there is more than one).
(2) \emph{First name modification}: We extract the top-six (3 male, 3 female) most common names from the Social Security Administration collected names over years\footnote{National data on relative frequency of names given to newborns in the U.S. assigned a social security number: \url{http://www.ssa.gov/oact/babynames}.} as artifacts. 
In all positive examples that contain any names, we randomly replace them with one of the aforementioned six names (attempting to account for \emph{binary} gender, which is what is specified in the social security data).
(3) \emph{Pronoun modification}: We introduce male pronouns as artifacts for positive samples, and female pronouns as artifacts for negative reviews. 
Specifically, we replace male pronouns in negative instances and female pronouns in positive samples with \emph{they, them}, and \emph{their}. 
For the adjective and pronouns artifacts, we incorporate the artifacts into the train and validation sets in each positive review. 
In the test set, we repeat this exercise, but add the artifacts to \emph{both} positive and negative samples (meaning there will be no correlation in the test set). 

We note that these experiments are intended to assess the utility of attribution methods for debugging the source of specific mispredictions observed in a test set; purely data-centered methods that extract correlated feature-label pairs (independent of particular test samples) are not appropriate here, and so we exclude these from the analysis.

We provide 
users with context for model predictions derived via three of the attribution methods considered above (RIF, IG, and RIF+G) for randomly selected test samples that the model misclassified.
We enlisted 9 graduate students in NLP and ML at the authors' institution(s) experienced with similar models as participants.
Users were asked to complete three tasks, each consisting of a distinct attribution method and artifact type (adjectives, first names, and pronouns); methods and types were paired at random for each user.
For each such pair, the user was shown 10 different reviews. 

Based on these examples, we ask users to identify: (1) \textit{The most probable artifacts,\footnote{We described artifacts to users as correlations between annotated sentiment of train reviews and the presence/absence of specific words in the review text.}} and, (2) \textit{the label aligned with each artifact}. 
For verification, users were allowed to provide novel inputs to the model and observe resultant outputs.
We 
recorded the number of model calls and the total engagement time to
evaluate efficiency (We provide a screenshot of our interface in the Appendix, Section \ref{ap:us}).

We report the accuracy with which users were able to correctly determine the artifact in Table \ref{tab:user-study}.
Users were better able to identify artifacts 
using $\metas$. 
Moreover, users spent the most amount of time and invoked the model more in $\metas$ case, which 
may be because inferring artifacts from influential training features requires more interaction with the model.
Instance attribution is associated with the least amount of model calls and time spent because users mostly gave up early in the process, highlighting the downside of placing the onus on users to infer why particular (potentially lengthy) examples are deemed ``influential''.


\begin{table}
\small
\centering
\begin{tabular}{lccccc}
\toprule
&\bf Acc&\bf Label-Acc&\bf \#Calls&\bf Time (m)\\
\midrule
RIF&3.7 & \bf 100.0 &\bf 6.4 &\bf 8.0\\
IG&31.6 & \bf 100.0 & 22.1& 8.2\\
RIF+G& \bf 47.0 & 94.5 & 28.6 & 10.1  \\
\bottomrule
\end{tabular}
\caption{We report: Average user accuracy (\textbf{Acc}) achieved, 
in terms of identifying inserted artifacts; How often users align artifacts with correct \textbf{labels}; The average number user interactions with the model (\textbf{\#Calls}), and; Average engagement \textbf{time} for each method.}  
\label{tab:user-study}
    \minipostspace{}
\end{table}
\section{Related Work}
\vspace{-0.5em}

\para{Artifact Discovery}
Previous studies approach the concerning affairs of artifacts by introducing datasets to facilitate investigating models' reliance on them~\citep{mccoy2019right}, analyzing existing artifacts and their effects on models \citep{gururangan2018annotation}, using instance attribution methods to surface artifacts and reduce model bias \citep{han2021influence,zylberajch2021hildif}, or use artifact detection as a metric to evaluate interpretability methods \citep{ross2020explaining}. 
To the best of our knowledge, only one previous work \citep{han2020explaining} set out to provide a methodical approach to artifact detection. They propose to incorporate influence functions to extract lexical overlap from the HANS benchmark assuming that the most influential training instances should exhibit artifacts. 
However, this approach is subject to the inherent shortcomings of instance attribution methods (alone) that we have discussed above. 
This work also assumed that the artifact sought was known \emph{a priori}.
Finally, \citet{gardner2021competency} investigate artifacts philosophically, theoretically analyzing spurious correlations in features.  

\para{Features of Training Instances} \citet{koh2017understanding} provided an approximation on training feature influence (i.e., the effect of perturbing individual training instance features on a prediction), 
and used this approximation in adversarial attack/defense scenarios. 
By contrast, here we have considered $\metas$ in the context of identifying artifacts, and introduced a broader set of such methods. 

\section{Conclusions}
\vspace{-.5em}
\emph{Artifacts}---here operationally defined as spurious correlations  
in labeled between features and targets that owe to incidental properties of data collection---can lead to misleadingly ``good'' performance on benchmark tasks, and to poor model generalization in practice.
Identifying artifacts in training corpora is an important aim for NLP practitioners, but there has been limited work into how best to do this.

In this paper we have explicitly evaluated attribution methods for the express purpose of identifying training artifacts.
Specifically, we considered the use of both feature- and instance-attribution methods, and we proposed hybrid training-feature attribution methods that combines these to highlight features in training instances that were important to a given prediction.
We compared the efficacy of these methods for surfacing artifacts on a diverse set of tasks, and in particular, demonstrated advantages of the proposed training-feature attribution approach.
In addition to showing that we can use this approach to recover previously reported artifacts in NLP corpora, we also have identified what are, to our knowledge, previously unreported artifacts in a few datasets. 
Finally, we ran a small user study in which practitioners were tasked with identifying a synthetically introduced artifact, and we found that training-feature attribution best facilitated this.
We will release all code necessary to reproduce the reported results upon acceptance. 

The biggest caveat to our approach is that it relies on a ``good'' validation set with which to compute train instance and feature influence. 
Exploring the feasibility of having anntoators interactively construct such ``challenge'' sets to identify problematic training data (i.e., artifacts) may constitute a promising avenue for future work.
All code necessary to reproduce the results reported in this paper is available at: \url{https://github.com/pouyapez/artifact_detection}.

\section*{Broader Impact Statement}
As large pre-trained language models 
are increasingly being deployed in the real world, 
there is an accompanying need to characterize potential failure modes of such models to avoid harms.
In particular, it is now widely appreciated that training such models over large corpora commonly introduces biases into model predictions, and other undesirable behaviors.
Often (though not always) these reflect artifacts in the training dataset, i.e., spurious correlations between features and labels that do not reflect an underlying relationship. 
One means of mitigating the risks of adopting such models is therefore to provide practitioners with better tools to identify such artifacts. 

In this work we have evaluated existing interpretability methods for purposes of artifact detection across several case studies, and we have introduced and evaluated new, hybrid $\meta$ methods for the same. 
Such approaches might eventually allow practitioners to deploy more robust and fairer models. 
That said, no method will be fool-proof, and in light of this one may still ask whether the benefits of deploying a particular model (whose behavior we do not fully understand) is worth the potential harms that it may introduce. 

\section*{Acknowledgements}

We would like to thank the anonymous reviewers for their feedback. 
Further, we also thank Matt Gardner, Daniel khashabi, Robert Logan, Dheeru Dua, Anthony Chen, Yasaman Razeghi and Kolby Nottingham for their useful comments. 
This work was sponsored in part by the Army Research Office (W911NF1810328), in part by the NSF grants \#IIS-1750978, \#IIS-2008956, \#IIS-2040989, and \#IIS-1901117, and a PhD fellowship gift from NEC Laboratories. 
The views expressed are those of the authors and do not reflect the policy of the funding agencies.

\clearpage
\bibliography{main}
\bibliographystyle{acl_natbib}

\clearpage
\section*{Appendix}
\appendix
\appendix


\section{Experimental Setup}
\vspace{-.5em}
\label{ap:es}
\paragraph{Datasets}
To investigate artifact detection, we conduct experiments on several common NLP benchmarks. 
We consider two benchmarks with previously known artifacts: (1) HANS dataset \citep{mccoy2019right}, which comprises 30k examples exhibiting previously identified NLI artifacts such as lexical overlap between hypotheses and premises. We randomly sampled 1000 instances from this benchmark as test data and use 10k randomly sampled instances from the Multi-Genre NLI (MNLI) dataset \citep{williams2017broad}, which contains 393k pairs of premise and hypothesis from 10 different genres, as training data. (2) We also use the 
IMDB binary sentiment classification corpus \citep{maas2011learning}, comprising 25k training and 25k testing instances. 
It has been shown in prior work \citep{ross2020explaining} that models tend to rely on the presence of ratings (range: 1 to 10) within IMDB review texts as artifacts. 

We have also reported novel (i.e., previously unreported) artifacts in several benchmarks. 
These include: (1) The DWMW17 dataset \citep{davidson2017automated} which is composed of 25K tweets labeled as \emph{hate speech}, \emph{offensive}, or \emph{non-toxic}; 
(2) BoolQ \citep{clark2019boolq}, a question answering dataset which contains 16k pairs of yes/no answers and corresponding passages.

\paragraph{Models} We adopt BERT \citep{devlin2019bert} with a linear model on top as a classifier and tune hyperparameters on validation data via grid search.
Specifically, tuned hyperparameters include the regularization parameter $\lambda=[10^{-1}, 10^{-2}, 10^{-3}]$; learning rate $\alpha=[10^{-3}, 10^{-4}, 10^{-5},10^{-6}]$; number of epochs $\in \{3,4,5,6,7,8\}$; and the batch size $\in \{8, 16\}$. 
Our final model accuracy on the benchmarks are as follows: \textit{IMDB:} 93.2\%, \textit{DWMW17:} 91.1\%, \textit{BoolQ:} 77.5\%.


\paragraph{Calculating the Gradient} To calculate gradients for individual tokens, we adopt a similar approach to \citet{atanasova2020diagnostic}, i.e., calculating the gradient of output (before the softmax), or instance attribution score with respect to the token embedding. 
We aggregate the resulting vector by taking an average; this has shown to be effective in prior work \citet{atanasova2020diagnostic} and provides a sense of positively and negatively influential tokens for model predictions (as compared to using $L2$ norm as an aggregating function).

\begin{figure*}[tb]
    \centering
    \includegraphics[width=.9\linewidth]{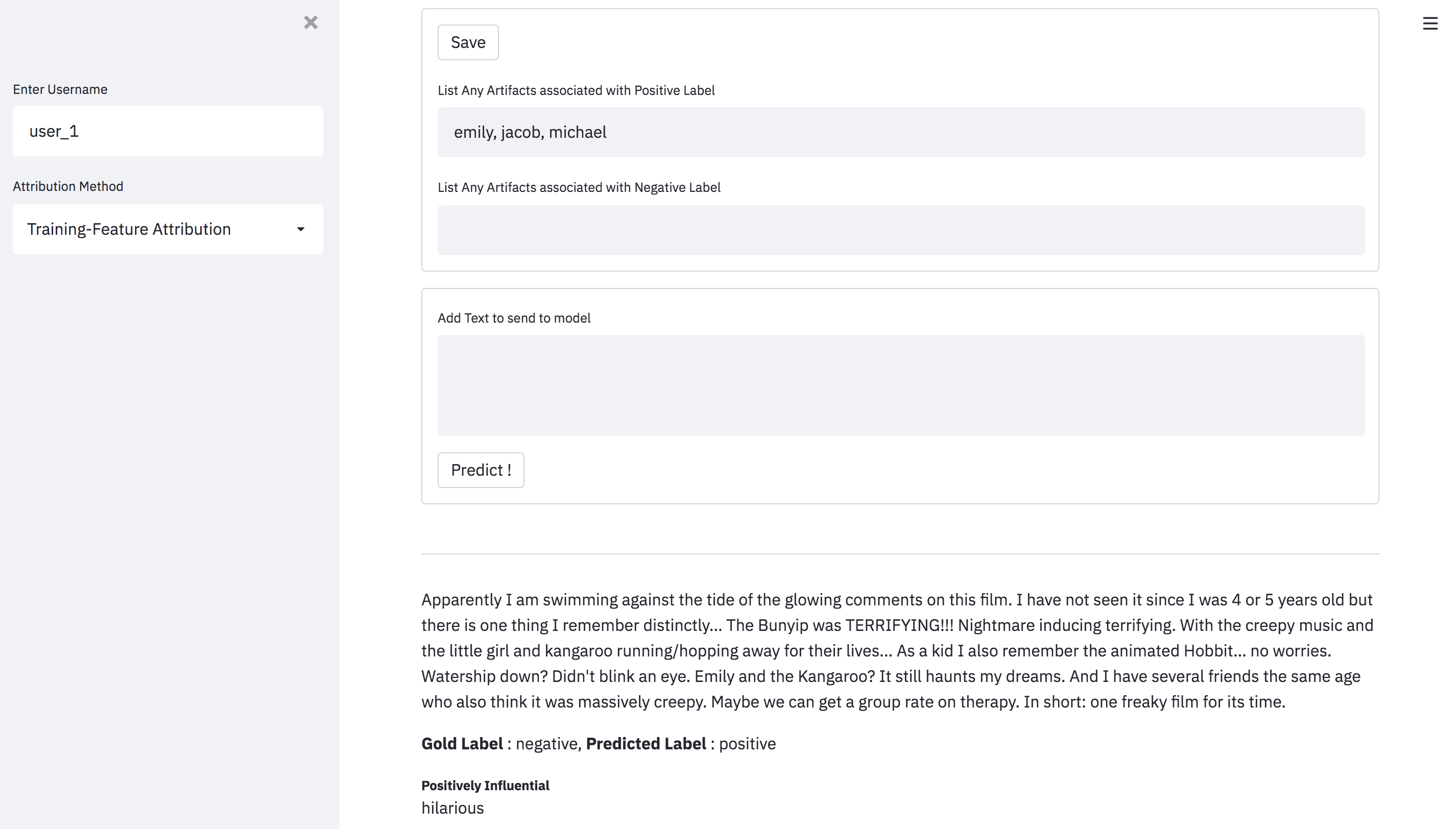}
    \caption{Screenshot of the user study's interface.}
    \label{fig:screen}
\end{figure*}

\section{User Study}
\label{ap:us}
The list of randomly sampled neutral adjectives, most popular names, and the pronouns used as artifacts are as follows: \emph{Adjectives} = [regular, cinematic, dramatic, bizarre ,artistic, mysterious], \emph{First-names} = [Jacob, Michael, Ethan, Emma, Isabella, Emily] and \emph{Pronouns} = [he, his, him, she, her]. 
We also provide a screenshot of the interface used in our user study in Figure \ref{fig:screen}.





\end{document}